\documentclass[fleqn,10pt]{wlscirep}
\usepackage[utf8]{inputenc}
\usepackage[T1]{fontenc}
\usepackage{booktabs}
\usepackage{multirow}
\usepackage{array}
\usepackage{xcolor}
\usepackage{amssymb}

\usepackage[hashEnumerators,smartEllipses]{markdown}

\usepackage{lineno}

\usepackage{tikz}
\usetikzlibrary{positioning,calc}
\usepackage{array}
\usepackage{tabularx}

\usepackage{pifont}

\usepackage{makecell}
\usepackage{pdfpages}
\usepackage{enumitem}

\usepackage[numbers]{natbib}

\title{FAIR² Drones: An AI-Ready Standard for Cross-Domain Wildlife Drone Datasets}

\author[1,*]{Jenna Kline}
\author[2]{Kilian Meier}
\author[3,4]{Vandita Shukla}
\author[5] {Edouard G. A. Rolland}
\author[6,7]{Elena Iannino}
\author[8]{Lucie Laporte-Devylder}
\author[4]{Constanza Andrea Molina Catricheo}
\author[6,7]{Blair Costelloe}
\author[1] {Elizabeth Campolongo}
\author[5] {Henrik S. Midtiby}
\author[9]{Devis Tuia}
\author[4]{Benjamin Risse}
\author[5]{Ulrik P.S. Lundquist}
\author[5]{Anders Lyhne Christensen}
\author[3]{Fabio Remondino}
\author[2]{Thomas Richardson}
\author[1]{Tanya Berger-Wolf}

\affil[1]{The Ohio State University, Department of Computer Science and Engineering, Columbus, OH, USA}

\affil[2]{School of Civil, Aerospace and Design Engineering, University of Bristol, Bristol, United Kingdom}

\affil[3]{3D Optical Metrology (3DOM), Fondazione Bruno Kessler (FBK), Trento, Italy}

\affil[4]{Computer Vision and Machine Learning Systems Group, Institute for Geoinformatics, University of Muenster, Muenster, Germany}

\affil[5]{Unmanned Aerial Systems Center, University of Southern Denmark, Odense, Denmark}

\affil[6] {Department of Collective Behavior, Max Planck Institute of Animal Behavior, Konstanz, Germany}
\affil[7]{Department of Biology, University of Konstanz, Konstanz, Germany}

\affil[8]{Department of Biology,  University of Southern Denmark, Odense, Denmark}

\affil[9]{Environmental Computational Science and Earth Observation laboratory, Ecole Polytechnique Fédérale de Lausanne}

\begin{abstract}
    Animal ecology data collection using drones represents a substantial investment of time, expertise, and financial resources. Yet most existing datasets serve only a single research community, limiting interdisciplinary reuse. We propose a unified drone dataset standard, \textbf{FAIR² Drones}, that bridges ecology, robotics, and computer vision by building on existing FAIR and AI-ready data frameworks while adding essential platform metadata and annotation specifications. Our standard enables datasets to simultaneously support ecological analysis, robotics algorithm development, and computer vision benchmarking. We provide open-source validation tools, reference implementations, and multimodal extensions linking drone imagery with complementary sensors such as camera traps, GPS, and acoustics. By standardizing metadata across disciplines, this framework maximizes the scientific return on investment for costly field deployments and accelerates cross-domain collaboration in environmental monitoring.
\end{abstract}

\begin{document}
\maketitle

\section{Introduction}
Collecting ecological data with drones for wildlife studies is both costly and logistically demanding~\cite{kline2025studying, pedrazzi2025advancing, Hughey2018challenges}. 
Deployments require specialized equipment, trained operators, and frequently, governmental or conservation permits~\cite{Maalouf2025Insights, lundquist_wilddrone_2026}. 
Ethical and safe operations also depend on close collaboration with local experts to ensure animal welfare~\cite{afridi2025impact}. 
Even short field campaigns can accumulate substantial expenses once travel, coordination, and post-processing are included~\cite{kline2025wildwing}. Given these investments, each drone dataset represents a significant scientific resource. Yet, most are designed for a single research purpose, often within one disciplinary context, which limits their broader reusability. At the same time, drones have become indispensable tools for wildlife research~\cite{pedrazzi2025advancing, Corcoran2021automated, Schad2023opportunities, Iglay2024Wildlife}. Drones provide safe, efficient access to remote habitats and can collect high-resolution imagery with minimal disturbance compared to ground-based methods. The recent progress of computer vision and machine learning now enables the rapid, automated analysis of such imagery~\cite{kline2025studying, Tuia2022perspectives}, from detecting animals~\cite{Corcoran2021automated} to identifying individuals~\cite{desai2022identification} and even inferring behavior~\cite{Ozogany2023fine, kline2025kabr}. This convergence of ecology, AI, and robotics presents new opportunities for interdisciplinary discovery but also new challenges in data management and interoperability.

Through our experience in the \textit{WildDroneEU} project~\cite{lundquist_wilddrone_2026}, which unites researchers in computer vision, robotics, and animal ecology, we identified a persistent obstacle: most drone imagery datasets are created for narrowly defined objectives and lack the contextual metadata required for reuse across domains. Computer vision datasets often omit ecological context, such as species behavior and habitat, while ecological datasets rarely include technical details crucial for robotics or autonomy research, such as flight trajectories, control modes, or camera calibration details. As a result, valuable data remain underutilized beyond their original purpose, constraining interdisciplinary collaboration and slowing progress.

FAIR principles--Findable, Accessible, Interoperable, and Reusable--guide open data sharing across disciplines~\cite{Barnas2020Standardized}. 
Ecology, robotics, and computer vision communities have established standards for sharing and evaluating datasets \cite{ellsaser_towards_2025, wieczorek2012darwin, Guralnick2018Humboldt, akhtar2024croissant}. Sharing datasets in a usable, standardized format promotes open science and more transparent and replicable research~\cite{gebru2021datasheets}. 
Publishing data allows peers to validate findings and encourages building on shared tools~\cite{yang2024navigating}. In additional to data and associated labels, metadata describing how data was collected and processed is essential for reproducibility and secondary use~\cite{Jenkins2024Reproducibility}. However, existing animal ecology drone datasets seldom meet baseline FAIR principles~\cite{Wilkinson2016FAIR}, let alone the extended FAIR$^{2}$ guidelines for AI-readiness \cite{Huerta2023FAIR}. Without standardized metadata, it becomes difficult to align datasets across communities or support machine-actionable interoperability. 

We propose that with modest additional effort during curation, researchers can greatly increase the utility of their data. A dataset collected for one goal, such as animal detection, could also advance robotics benchmarking, ecological analysis, or behavior modeling if accompanied by comprehensive metadata and standardized annotations. Some drone datasets collected for animal ecology studies meet standards from one or more fields, but gaps remain.  We aim to bridge and extend these existing frameworks to accommodate cross-domain drone data. FAIR$^{2}$ extends the broadly-used FAIR framework for AI-ready datasets~\cite{Huerta2023FAIR}, emphasizing structured metadata, computational accessibility, provenance, and bias transparency. We adopt FAIR$^{2}$ as the conceptual basis for our standard, aligning it with the Darwin Core ecological framework~\cite{wieczorek2012darwin} to represent species observations and sampling events.

We propose the \textbf{FAIR² Drones} dataset standard, a framework tailored for drone-based wildlife monitoring. 
The FAIR² Drones standard builds on existing foundations from across disciplines, including the Darwin Core Humboldt Extension~\cite{wieczorek2012darwin} and FAIR$^{2}$~\cite{Huerta2023FAIR} to define a unified structure for describing platform specifications, sensor details, autonomy modes, and ecological context. Meeting both FAIR$^{2}$  and Darwin Core compliance forms the minimal interoperability threshold for drone-based ecological data. By harmonizing these elements, drone datasets can become AI-ready, interoperable, and recontextualizable across computer vision, robotics, and ecology. This will enable more efficient data reuse, foster collaboration between disciplines, and accelerate scientific discovery.
\section{Background}
\begin{table*}[t]
\centering
\caption{Comparison of existing animal monitoring datasets across disciplinary requirements. Checkmarks (\checkmark) indicate comprehensive coverage, warning symbols (\textcolor{orange}{$\Delta$}) indicate partial coverage, and crosses (×) indicate absence of required metadata. 
\footnotesize{Definitions: Uncrewed Aerial Vehicle (\textbf{UAV}); Taxonomic Databases Working Group (\textbf{TDWG}) \cite{TDWG2024Biodiversity}; Autonomous Underwater Vehicle (\textbf{AUV}); Remotely Operated Vehicle (\textbf{ROV}).}
}

\label{tab:dataset_comparison}
\small
\begin{tabular}{@{}p{3cm}p{1.5cm}p{2cm}p{2cm}p{2.5cm}p{1cm}p{3cm}@{}}
\toprule
\textbf{Dataset} &
\textbf{Platform} &
\textbf{Ecological Metadata} &
\textbf{Telemetry} &
\textbf{Computer Vision Annotations} &
\textbf{Multi-modal} &
\textbf{Limitation} \\
\midrule

KABR \cite{kline2025kabr} &
UAV &
\checkmark Full &
\checkmark Full &
\checkmark Multi-task &
× &
Not TDWG-aligned \\

\addlinespace[0.1cm]

BuckTales \cite{naik2024bucktales} &
Multi-UAV &
\checkmark Behavioral &
\textcolor{orange}{$\Delta$} Minimal &
\checkmark Tracking / Re-ID &
× &
No standardized metadata format \\

\addlinespace[0.1cm]

MMLA \cite{kline2025mmla} &
UAV &
\textcolor{orange}{$\Delta$} Species only &
\textcolor{orange}{$\Delta$} Partial &
\checkmark Detection &
× &
Missing ecological context \& telemetry \\

\addlinespace[0.1cm]

Galeda monkeys \& African ungulates \cite{Koger2023Quantifying} &
UAV &
\checkmark Behavioral ecology &
\checkmark Full &
\checkmark Pose \& tracks &
× &
Not AI-ready \\

\addlinespace[0.1cm]

WAID \cite{mou2023waid} &
UAV &
× None &
× None &
\checkmark Detection only &
× &
Missing platform information \& ecological context \\

\addlinespace[0.1cm]

Marine Tracking \cite{cai2023semisupervised} &
AUV/ROV &
\checkmark Species focus &
\textcolor{orange}{$\Delta$} Variable &
\checkmark Detection &
× &
No robotics metadata \\

\addlinespace[0.15cm]

\textbf{FAIR² Drones} &
\textbf{UAV/AUV} &
\checkmark \textbf{Darwin Core} &
\checkmark \textbf{Full specs} &
\checkmark \textbf{Multiple formats} &
\checkmark &
\textbf{—} \\

\bottomrule

\end{tabular}
\end{table*}

\subsection{Existing animal ecology drone datasets and standards}
Existing animal monitoring datasets span diverse platforms and research objectives but vary widely in how comprehensively they capture metadata required for cross-domain reuse. Animal ecology drone datasets are inherently heterogeneous: they may include photos or videos or both, and can include annotations for a range of computer vision tasks such as detection, tracking, or behavior analysis. Depending on the ecological research question, datasets may include context such as habitat characterization, species, life stage, and behavior observations. Ecological datasets may also focus on landscape maps to understand how animals interact with their environment \cite{Koger2023Quantifying}. Further, collecting data for ecological insight typically requires different sampling strategies than collecting data for computer vision or robotics benchmarking \cite{kline2025studying}.

As shown in Table \ref{tab:dataset_comparison}, most drone-based datasets provide rich computer-vision annotations but lack standardized ecological or platform metadata. KABR~\cite{kline2025kabr} offers the most complete coverage across ecology and computer vision but is not aligned with community biodiversity standards such as Darwin Core. The BuckTales~\cite{naik2024bucktales} and Multi-species, Multi-location, Low Altitude (MMLA) drone~\cite{kline2025mmla} datasets include behavioral and detection-level labels, respectively, yet provide only minimal information about flight parameters or sensor specifications, limiting reuse for robotics benchmarking. Datasets such as the Gelada monkeys and African ungulates collection demonstrate high ecological detail and pose tracking capability but do not include detailed telemetry logs~\cite{Koger2023Quantifying}, while the Wildlife Aerial Images from Drone (WAID) dataset~\cite{mou2023waid} and Marine Tracking~\cite{cai2023semisupervised} contain useful imagery but lack contextual or telemetry data. Collectively, these examples underscore that even well-curated drone datasets are often single-purpose, optimized for one community but difficult to integrate across others.

Barnas et al.~\cite{Barnas2020Standardized} established standards for reporting when using drones for wildlife research. This includes key metadata to document the full life cycle of a drone mission from design through data analysis. Their structured reporting protocol includes: (1) project overview, (2) platform specifications, (3) payload and sensor details, (4) flight planning and operations, (5) data post-processing, and (6) permits, regulations, and training. These components comprehensively document the full life cycle of a drone mission from design through data analysis. Reporting on permits and regulations are particularly critical for complex missions such as beyond-visual-line-of-sight (BVLOS), night missions, and missions conducted at wildlife sanctuaries~\cite{Maalouf2025Insights}. For reproducibility and transparency, reporting operational and mission-level metadata such as flight trajectories, autonomy modes, environmental conditions, mission objectives, temporal information, and endurance constraints, all of which influence data quality and operational safety, are critical.

\subsection{Heterogeneous data standards for ecology}
Biodiversity data originate from a wide variety of sources, stored in diverse formats across different platforms. A crucial step toward understanding global biodiversity patterns is to provide a standardized framework that integrates these heterogeneous data to improve interoperability. Central to this effort is the definition of shared terminology. 
The Darwin Core standard was developed for this purpose, providing a stable, community-driven vocabulary for publishing and integrating biodiversity information~\cite{wieczorek2012darwin}. Darwin Core organizes biodiversity data around two central entities: the \textit{Event}, which captures the context of data collection (how, when, and where), and the \textit{Occurrence}, which records what was observed. We provide key metadata for \textit{Events} and \textit{Occurrences} in Table \ref{tab:darwin_core_fields}. These classes offer a robust foundation for representing ecological context in drone imagery and can be extended to describe aerial sampling events and detected organisms. These fields are widely adopted by major biodiversity repositories such as Global Biodiversity Information Facility (GBIF) and the Ocean Biodiversity Information System (OBIS), providing a natural foundation for integrating drone-derived observations into existing ecological data infrastructures. The Humboldt Extension provides a standardized foundation for ecological and sampling metadata~\cite{TDWG2024Humboldt}.

\begin{table}[h!]
\centering
\caption{\textbf{Core Darwin Core metadata fields for biodiversity observations.} 
Event metadata describe the data collection context, while Occurrence metadata describe the observed organisms.}
\begin{tabular}{p{4cm} p{12cm}}
\toprule
\textbf{Category} & \textbf{Key Metadata Fields and Descriptions} \\
\midrule
\textbf{Event} (data collection context) & 
\textit{eventID} – unique identifier for each sampling event; 
\textit{eventDate}, \textit{eventTime} – date and time of data capture; 
\textit{decimalLatitude}, \textit{decimalLongitude}, \textit{coordinateUncertaintyInMeters} – geolocation and precision; 
\textit{habitat} – description of environment; 
\textit{samplingProtocol} – method or technology used (e.g., drone survey, camera trap); 
\textit{sampleSizeValue}, \textit{sampleSizeUnit} – spatial or temporal extent of sampling; 
\textit{samplingEffort} – qualitative or quantitative measure of effort expended. \\[0.75em]

\textbf{Occurrence} (observed organisms) &
\textit{occurrenceID} – unique identifier for each observation; 
\textit{scientificName}, \textit{taxonRank}, \textit{kingdom–species} – taxonomic classification; 
\textit{individualCount} – number of individuals observed; 
\textit{lifeStage}, \textit{sex}, \textit{behavior} – biological traits and activity; 
\textit{occurrenceRemarks} – additional notes on observation or detection context. \\
\bottomrule
\end{tabular}
\label{tab:darwin_core_fields}
\end{table}

\subsection{FAIR and AI-readiness}
Recent work on FAIR² data principles extends the original FAIR framework to explicitly support AI-ready datasets~\cite{Huerta2023FAIR}. AI-ready datasets must not only reusable by humans but actionable by machines. FAIR² emphasizes machine-readable metadata, semantic consistency, and transparency about data provenance, quality, and bias which are critical elements for training robust, reproducible AI models. In the context of animal monitoring, most existing datasets lack the metadata richness and cross-domain context necessary for AI applications beyond their original scope. Ecological frameworks, such as Darwin Core, capture rich taxonomic and occurrence metadata but omit details about sensor payloads, flight control, or autonomy. Conversely, computer-vision standards like COCO and PASCAL VOC define robust annotation schemas for object detection and segmentation but include no ecological or collection context.
Robotics standards such as  the Motion Imagery Standards Board (MISB) for Full Motion Video Add-in's~\cite{MISB} and the MAVLink drone messaging protocol~\cite{koubaa2019micro} capture sensor interfaces yet lack biological or behavioral descriptors.
This misalignment across domains prevents the seamless fusion of data needed for multi-modal ecological studies and cross-disciplinary AI development.

Our proposed standard operationalizes FAIR² by embedding requirements for dataset documentation, annotation provenance, versioning, and multi-modal synchronization. Preparing dataset cards following this schema ensures that each dataset is self-describing and machine-interpretable, detailing ecological context, sensor configuration, annotation methodology, and data quality metrics. These cards act as a bridge between FAIR² principles and practical dataset publication, enabling automated validation, benchmarking, and model training across ecology, robotics, and computer vision. In doing so, our framework advances both the Humboldt-compatible ecological data ecosystem and the growing movement toward FAIR², AI-ready scientific data. 
\section{Cross-Disciplinary Metadata Requirements}
\label{requirements}

\begin{table*}[t]
\centering
\caption{Interdisciplinary metadata requirements for ecological monitoring with autonomous systems. Domains span ecology, robotics, computer vision, multimodal sensing, and cross-cutting standards.}
\label{tab:requirements}
\small
\begin{tabular}{@{}p{2.4cm}p{3.2cm}p{11cm}@{}}
\toprule
\textbf{Domain} & \textbf{Metadata Category} & \textbf{Key Requirements} \\
\midrule

\multirowcell{5}[0pt][l]{\textbf{Ecology}} 
& Species identification & Include both scientific and common names for all observed taxa. \\
& Behavioral context & Ethograms, activity budgets, or behavioral state annotations linking visual data to context. \\
& Environmental metadata & Habitat type, vegetation cover, weather conditions, and season of observation. \\
& Sampling methodology & Protocol, temporal coverage, and potential sampling biases. \\
& Darwin Core / Humboldt & Compliance with biodiversity standards for interoperability and citation. \\

\midrule

\multirowcell{5}[0pt][l]{\textbf{Robotics}} 
& Platform specifications & UAV model, payload configuration, sensor suite, and autonomy mode. \\
& Telemetry data & GPS, IMU, altitude, speed, orientation, and battery levels. \\
& Mission planning & Waypoints, coverage maps, control modes, and flight duration. \\
& Sensor calibration & Temporal and spatial synchronization, photogrammetric sensor calibration. \\
& Regulations & Flight permits, safety protocols, and local collaboration documentation. \\

\midrule

\multirowcell{4}[0pt][l]{\textbf{Computer Vision}} 
& Annotation format & COCO, YOLO, or Pascal VOC-compliant annotation structures. \\
& Data partitioning & Defined train/validation/test splits with class distribution statistics. \\
& Quality metrics & Inter-annotator agreement, label confidence, and occlusion or truncation flags. \\
& Difficulty metadata & Indicators for visibility, crowding, background clutter, or environmental complexity. \\

\midrule

\multirowcell{4}[0pt][l]{\textbf{Synchronization}} 
& Temporal & Timestamps, frame alignment, correction for clock drift, timezone and frequency. \\
& Spatial calibration & Coordinate system definitions and transformation matrices. \\
& Sensor-specific metadata & Thermal calibration, LiDAR beam pattern, acoustic frequency range. \\
& Discoverability & Persistent identifiers (PIDs) and machine-readable metadata to support data reuse. \\

\midrule

\multirowcell{4}[0pt][l]{\textbf{Cross-Cutting}} 
& Provenance \& version & Full traceability from raw to processed data products. \\
& Quality control & Inclusion of uncertainty estimates, validation logs, and quality flags. \\
& Coordinate systems & Consistent spatial reference frameworks across datasets and domains. \\
& FAIR$^2$ compliance & Findability, accessibility, interoperability, and reusability for both humans and machines. \\

\bottomrule
\end{tabular}
\end{table*}

Each research domain brings distinct requirements for dataset documentation (Table \ref{tab:requirements}). Rather than proposing entirely new standards, we identify how existing frameworks can be extended and integrated to serve the interdisciplinary needs of drone-based wildlife monitoring.

\subsection{Ecological Requirements}
For animal ecology studies, drones capture photos and videos of wildlife in their natural habitats, documenting not only the animals themselves but also behaviors, intraspecific interactions, and habitat features. Beyond imagery, these missions generate observational data requiring expert interpretation, such as social and environmental context. Species identification should include both scientific and common names, with taxonomic validation against authoritative databases. For behavioral studies, ethograms and activity budgets require documentation of social context (e.g., group composition, proximity to conspecifics) and environmental conditions including habitat type, weather, season, and time of day. Sampling effort details, including coverage area, temporal extent, and potential biases, are essential for statistical inference. Research permits and ethical approvals should also be documented. These requirements align closely with Darwin Core \cite{wieczorek2012darwin} and the Humboldt Extension for ecological inventories \cite{TDWG2024Humboldt}, providing an established foundation for biodiversity data interoperability.

\subsection{Robotics Requirements}
Robotics datasets center on drone telemetry: GNSS position, IMU readings, altitude, speed, orientation, camera and gimbal position, and battery status captured during operation. This telemetry enables post-hoc analysis of flight performance and refinement of automated collection protocols.
Previous works have established metadata requirements for FAIR drone datasets, including documentation of the sensor payload, spatial and temporal coverage, data processing workflows, and provenance \cite{ellsaser_towards_2025}.
Platform specifications should include drone model, sensor suite, and payload configuration. 
Sensor calibration measurements, when available, support photogrammetric analysis and spatial accuracy assessment. Control modes and autonomy levels, such as obstacle avoidance, automated flight planning, or manual operation, should be documented to contextualize data collection conditions. Flight Logs provided by open-source flight controller software such as ArduPilot or PX4 are easily retrievable. Moreover, for most commercial platforms (DJI, Parrot, etc), flight logs can be exported in a consistent format with popular flight data platforms, such as AirData~\cite{airdata}. 

\subsection{Computer Vision Requirements}
Drone imagery provides rich training data for computer vision models that can accelerate ecological analysis~\cite{kline2025studying}. 
However, pretrained models rarely generalize to underrepresented or regionally distinct species, necessitating annotated data for fine-tuning~\cite{kline2025mmla,stevens2025mind}.
Annotations should follow standardized formats (COCO, YOLO, Pascal VOC, or MOT for tracking tasks) to ensure compatibility with existing tools, such as the Computer Vision Annotation Tool (CVAT)~\cite{cvat2023} or Kenyan Animal Behavior Tools (kabr-tools) \cite{kline2025kabr}.
Further, datasets should aim for compatibility with the Croissant  metadata format for ML-ready datasets\cite{akhtar2024croissant}, which enables the data to be reused and shared across platforms/
Dataset documentation should include train/validation/test splits and class distribution statistics \cite{huggingface2022datasets, pushkarna2022data}. For manual annotations, the methodology, annotation tool, and quality metrics such as inter-annotator agreement should be reported . Flags indicating occlusion, truncation, or annotation difficulty help users assess label reliability and filter samples appropriately.

\subsection{Synchronization and Multi-Modal Requirements}
Integrating data from multiple sensors and linking drone imagery to complementary data sources, such as GNSS collars, camera traps, acoustic recorders, requires careful synchronization across sensor modalities, temporal domains, and spatial reference frames. While existing standards like Darwin Core provide robust frameworks for spatial metadata and MISB address motion imagery metadata, critical gaps remain in documenting temporal synchronization protocols and sensor-specific calibration parameters essential for drone-based ecological monitoring.

\subsubsection{Temporal synchronization}
Temporal synchronization requires precise timestamps and, where applicable, correction for clock drift between sensors.
Darwin Core provides \textit{eventDate} and \textit{eventTime} fields for temporal documentation, which work well for traditional ecological sampling where a single timestamp captures the sampling event. However, drone missions generate heterogeneous temporal data streams that challenge this model, e.g. frame-based telemetry, high-frequency autopilot telemetry, irregular sampling of AI annotation. 
These data streams must be temporally aligned to enable frame-level correspondence between imagery, telemetry, and ecological observations, yet existing standards provide no vocabulary for documenting the sampling rates of different streams or the methods used to align them.

Timestamp heterogeneity compounds the challenge of temporal synchronization. 
Telemetry files may use UTC, local time, GNSS time, platform-specific epoch, or video-based timestamps. Darwin Core's temporal fields do not document which time standard is used or how conversions were performed. Clock drift between independent sensors is common, and researchers often employ manual synchronization protocols like visual markers, but there is no standardized field to document these methods or the achieved temporal accuracy. 

\subsubsection{Spatial synchronization}

Spatial calibration documentation should include coordinate system definitions and transformation matrices relating sensor reference frames.
Darwin Core fields capture the essential components of spatial documentation: where an observation occurred, which coordinate reference system was used, and the uncertainty inherent in that measurement. For most drone applications, these fields adequately document spatial information. Consumer-grade drones typically use WGS84 (EPSG:4326) as their geodetic datum, with GNSS horizontal accuracy on the order of 1-5 meters depending on satellite visibility and atmospheric conditions. This accuracy is sufficient for many ecological analyses, and Darwin Core's existing fields document it appropriately. In order to perform georectification tasks, such as estimating the locations of animals on the ground, camera calibration and ground control points are required~\cite{Koger2023Quantifying,Meier2025}.

Spatial privacy constraints create documentation challenges not addressed by existing standards. When monitoring endangered species, conducting research in protected areas with restricted access, or operating in regions where wildlife poaching is a concern, publishing absolute GNSS coordinates poses risks. Researchers may need to encrypt or systematically offset coordinates to protect sensitive locations. However, many analyses require knowing the relative spatial relationships within a dataset even if absolute positions are obscured. For example, calculating home range sizes or movement distances requires accurate inter-point distances but not necessarily true geographic coordinates. Existing biodiversity standards do not provide guidance on privacy-preserving spatial documentation protocols or fields to indicate that coordinates have been intentionally obscured while maintaining internal consistency. Even with exact GNSS location obscured, drone imagery can reveal approximate location, especially if known landmarks are captured. Researchers should work closely with conservation experts and wildlife management partners to ensure data released does not pose risks to wildlife welfare.
 
\subsubsection{Sensor Calibration and Multi-Modal Metadata}
Drones carrying multiple sensor payloads introduce additional calibration requirements.
Thermal sensors require radiometric calibration and emissivity settings, LiDAR systems need beam pattern specifications, stereo cameras require baseline distances and calibration matrices, and multi spectral sensors need spectral response curves. 
These sensor-specific parameters are absent from Darwin Core and computer vision annotation formats. Additionally, coordinate frame transformations between sensors, which is critical for data fusion, are rarely documented, preventing accurate integration of multi-sensor datasets.
The MISB has established comprehensive metadata schemas for motion imagery that would address many drone synchronization requirements \cite{MISB}. However, MISB adoption in ecological research is uncommon due to the technical complexity and its focus on defense applications rather than scientific reproducibility needs.

\subsubsection{Cross-Modal Data Linkage}
Drone observations often complement other ecological monitoring methods including GNSS collars, camera traps, acoustic recorders, and expert field observations. Darwin Core's \textit{associatedMedia} field can link occurrences to related resources through uniform resource identifiers (URI), like URL links, but it lacks specific conventions for documenting persistent identifiers to complementary datasets. Without consistent conventions for structuring these references, discovering cross-modal linkages requires examining dataset-specific documentation rather than parsing standardized metadata fields, hindering integration into broader multi modal monitoring networks.

\subsection{Cross-Cutting Requirements}

\begin{figure}[t]
    \centering
    \includegraphics[width=0.6\linewidth]{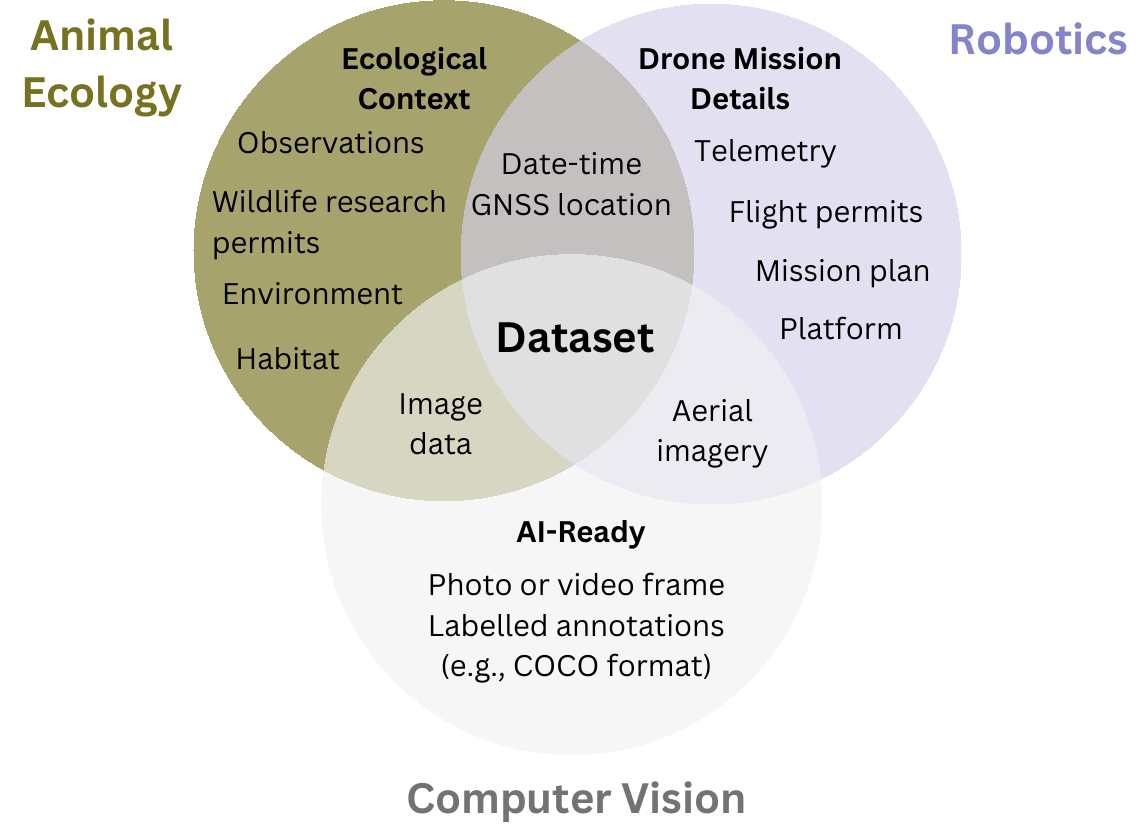}
    \caption{Data and metadata requirements for animal ecology, robotics, and computer vision research.}
    \label{fig:venn}
\end{figure}

Several requirements span all domains. Data provenance and versioning ensure traceability from raw acquisitions through processed products, supporting reproducibility and error diagnosis. Quality control documentation, including validation logs, uncertainty estimates, and quality flags, enables users to assess fitness for purpose. Consistent coordinate reference systems across datasets and domains facilitate spatial integration and comparison. Finally, adherence to FAIR principles (Findable, Accessible, Interoperable, Reusable) ensures datasets remain discoverable and valuable to both human researchers and automated systems~\cite{Wilkinson2016FAIR}. We extend this consideration to FAIR compliance, emphasizing machine-readability alongside human interpretability~\cite{Huerta2023FAIR}.
\section{FAIR² Drones Dataset Core Schema and Metadata}
\label{schema}
We demonstrate the practical utility and flexibility of the FAIR² Drone Data Standard through two complementary approaches: (1) detailed dataset cards that operationalize the standard's requirements 
, and (2) reference implementations retrofitting existing datasets to demonstrate backward compatibility and conversion pathways.

\subsection{Proposed Standard}
\label{standard}
Our proposed FAIR² Drone data standard was designed through consultation across ecology, robotics, and computer vision communities. Each domain brings distinct requirements reflecting its methodological traditions and research priorities, summarized in Table~\ref{tab:requirements}.
For the computer vision community, the value of a dataset lies in the quality, consistency, and transparency of its annotations. Benchmarks must follow standard formats and report data characteristics relevant to model training and evaluation.
Robotic requirements include platform specifications, telemetry, mission planning, control modes, calibration, and permissions. Robotics researchers require precise operational metadata to evaluate navigation, perception, and autonomy performance. These data support reproducibility and benchmarking of drone algorithms.

\begin{figure}
    \centering
\includegraphics[width=0.65\linewidth]{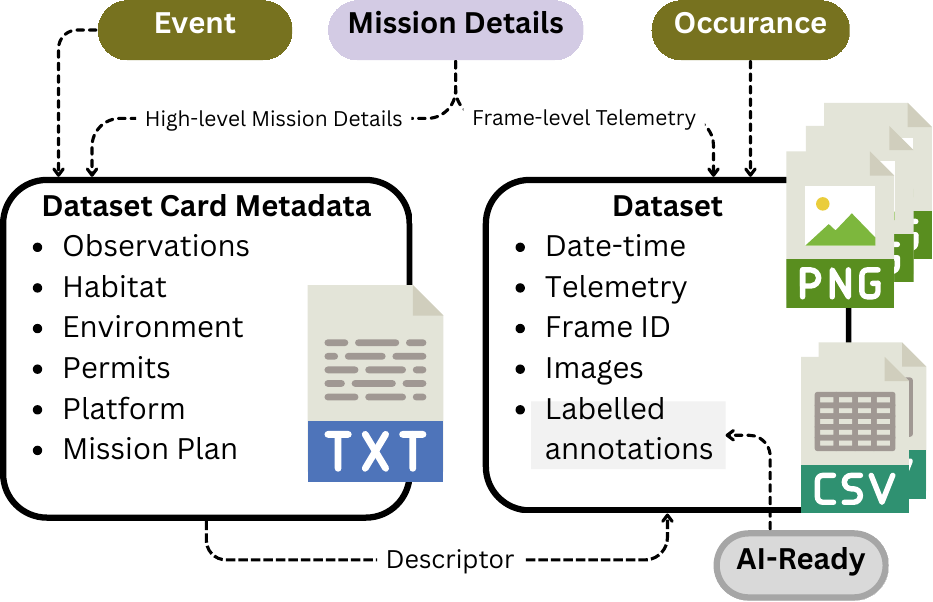}
    \caption{We propose standardized dataset cards as a bridge between the Humboldt ecological metadata framework and FAIR² (AI-ready) data principles. Our proposed dataset card schema integrates standard platform, sensor, and annotation metadata reporting from robotics and computer vision, encoded in machine-readable formats. Each dataset card maps relevant Darwin Core and Humboldt terms to AI-relevant attributes (e.g., synchronization, calibration, class balance, annotation provenance), ensuring data are both semantically rich and computationally actionable. By aligning with FAIR², this framework makes wildlife monitoring datasets findable, accessible, interoperable, reusable, recontextualizable, and actionable, enabling automated discovery and reuse across disciplines while maintaining ecological validity.}
    \label{fig:proposed}
\end{figure}

\subsection{FAIR² Dataset Card Template}

We implement FAIR² requirements through a \href{https://imageomics.github.io/fair_drones/}{Dataset Card Template} (detailed in Table~\ref{tab:fair2_structure}) that researchers complete using a familiar markdown format.
Dataset cards provide standardized documentation of dataset composition, methodology, and provenance for transparent and responsible data use \cite{gebru2021datasheets, pushkarna2022data}. Dataset cards have been widely adopted across AI-research communities \cite{yang2024navigating, huggingface2022datasets}. Our template extends the Imageomics dataset card \cite{Campolongo2025Collaborative, Campolongo_Imageomics_Guide_2025} with structured metadata.

\label{datasetcard}
\begin{table*}[t]
\centering

\caption{FAIR² Drone Datacard Structure }
\label{tab:fair2_structure}

\begin{tabular}{p{0.2cm} p{3.5cm} p{9.3cm} p{3cm}}
\toprule
\textbf{\#} & \textbf{Module} & \textbf{Contents} & \textbf{Purpose} \\
\midrule
1 & Dataset Description
& Persistent identifier (DOI), dataset title and version, curators, homepage, repository, and paper citation.
& Findability \\

2 & Supported Tasks
& Computer vision tasks (detection, tracking, segmentation), ecological applications, and robotics benchmarks with baseline metrics.
& Task Specification \\

3 & Dataset Structure
& Directory organization, file formats, naming conventions, Darwin Core Event/Occurrence records, and data field schemas.
& Interoperability \\

4 & Platform Specifications
& Robotic platform description, hardware components, autonomy modes, payload capacity, and navigation systems.
& Robotics Reproducibility \\

5 & Sensor Specifications
& Sensor models, calibration procedures, synchronization methods, spectral characteristics, and resolution specifications.
& Data Provenance \\

6 & Mission Parameters
& Flight altitude, speed, sampling protocol, environmental conditions, telemetry availability, and permits/ethics approvals.
& Operational Context \\

7 & Dataset Creation
& Data collection team, curation rationale, annotation process, quality control, and annotator demographics.
& Provenance \& Ethics \\

8 & Dataset Characterization
& Species distribution, class balance, temporal/spatial coverage, detection difficulty metrics, and known biases.
& Transparency \\

9 & Multimodal Linkages
& Associated datasets (camera traps, GPS collars, acoustics), temporal/spatial alignment methods, and cross-modal synchronization.
& Cross-Domain Integration \\

\bottomrule
\end{tabular}

\end{table*}

The template captures: (1)~\textbf{Platform}—defining the drone’s hardware, payload, and autonomy configuration; (2)~\textbf{Mission Context}—flight conditions, environmental variables, and sampling protocols; (3)~\textbf{Sensor Parameters}—calibration details and multi-sensor synchronization; (4)~\textbf{Annotation Schema}—harmonized species labels, detection, tracking, and/or behavior annotations, and quality metrics; (5)~\textbf{Cross-Modal Linkage}—synchronization with complementary sensing modalities (GNSS, acoustics, and camera traps); and (6)~\textbf{Provenance and Ethics}—data collection personnel, permitting documentation, and privacy protocols. Together, these metadata extensions promote AI-readiness, interoperability, and long-term reusability across ecology, robotics, and computer vision communities.

Researchers complete structured prompts for metadata fields aligned with Darwin Core's Event and Occurrence model, including \textit{scientificName, behavior, habitat, samplingProtocol}, and \textit{samplingEffort}. For AI-readiness, the template captures \textit{annotationQuality} metrics (inter-annotator agreement, validation procedures), \textit{classBalance} statistics, and \textit{difficultyScore} distributions to enable transparent model evaluation. Robotics applications benefit from control mode and autonomy-level descriptors that support reproducible benchmarking across field sites.

This design provides flexibility across research applications. Detection-only datasets complete platform, mission, and detection annotation modules. Multi-task datasets incorporating tracking and behavior fill all annotation subsections. Minimally-annotated robotics benchmarks can omit Darwin Core species fields while maintaining platform and mission metadata. Multimodal datasets expand cross-modal linkage sections to include acoustic, thermal, or GNSS collar synchronization. This tiered approach allows researchers to achieve baseline compliance in more quickly using semi-automated templates, while encouraging comprehensive documentation for richer datasets.

\section{Discussion and Future Work}

The FAIR² Drone data standard offers a practical, extensible foundation bridging ecology, robotics, and AI. By aligning Darwin Core, Humboldt Extension, and machine learning metadata conventions, we enable interoperability between ecological observation and computational modeling. Modern ecological monitoring increasingly relies on multimodal sensing systems that combine visual, acoustic, thermal, and environmental data streams to capture more complete understanding of ecosystems \cite{kline2025smartwilds}. Multimodal studies, such as MammAlps \cite{gabeff2025mammalps} and SmartWilds \cite{kline2025smartwilds},  demonstrate the value of coupling camera traps with passive acoustics to infer animal presence and behavior, while databases like MoveBank exemplify the impact of interoperability \cite{Kays2022Movebank}. Datasets within MoveBank utilize standardized formats, metadata vocabularies, and APIs that have transformed animal movement research through global-scale synthesis. 
Drones are emerging as the next essential component of these multimodal networks \cite{kline2025smartwilds}. 
By providing fine-grained spatial, temporal, and behavioral information, drones complement static sensors and can fill observational gaps in both space and time \cite{Tuia2022perspectives, besson2022towards}.
However, to function as interoperable nodes within these distributed systems, drone data must follow common standards that allow integration across sensing modalities and scientific domains. Without such standardization, multimodal datasets remain fragmented. Even when RGB and thermal cameras are co-mounted on a single drone, inconsistent metadata conventions prevent their outputs from being temporally aligned or spatially fused. The same issue arises when combining aerial imagery with ground-based camera traps, bioacoustic arrays, or water-quality sensors—interoperability hinges on shared schema and consistent metadata.

The next generation of ecological AI models will depend on multimodal training data. Ecological use cases include using RGB and thermal imagery for day-night detection, linking acoustic events with visual tracks to confirm species presence, or combining environmental and positional data to model animal-habitat interactions. 
A standardized metadata framework is what makes such fusion possible: aligning datasets through shared conventions for timestamps, spatial coordinates, and sensor specifications. 
When multimodal datasets are aligned and contextually interpretable, researchers can train generalizable AI models capable of learning relationships between modalities, environments, and species behaviors. These models, in turn, will enable adaptive autonomous systems that respond intelligently to changing ecological conditions.

\subsection{Limitations and Adoption Challenges}

\textbf{Standards proliferation and community buy-in}. The most significant challenge facing FAIR² is persuading researchers to adopt it when they may already be using established workflows or when multiple competing standards exist within their subdiscipline. We are acutely aware that proposing a new standard risks adding to fragmentation rather than solving it. However, FAIR² extends proven foundations (Darwin Core \cite{wieczorek2012darwin}, dataset card standards \cite{gebru2021datasheets}) rather than replacing them, reducing adoption barriers by building on vocabularies that repositories already support. Nevertheless, researchers face genuine barriers including time constraints, lack of training in metadata practices, and uncertain return on investment. Our tiered compliance structure addresses time concerns, but raises questions about what constitutes \emph{sufficient} metadata for different research communities.

\noindent\textbf{Sensitive species and spatial privacy.} Drone data inherently captures fine-grained location information, presenting concerns for endangered species monitoring. Publishing precise GNSS coordinates could enable poaching, while obscuring locations reduces data utility for spatial ecology. Our standard includes provisions for spatial generalization, but different species and regions require different privacy thresholds. Even without exact coordinates, metadata patterns (flight schedules, repeated area coverage) might betray sensitive locations. We recommend researchers work with local conservation authorities to determine appropriate obfuscation levels.

\noindent\textbf{Technological drift and maintenance.} Drone hardware and software evolve rapidly, with new sensors and data formats emerging constantly. Our standard must evolve alongside this technology or risk obsolescence. This raises critical questions of governance: who maintains FAIR² long-term, and how are extensions coordinated? We have designed FAIR² to be extensible through optional fields, but this flexibility could hinder interoperability if extensions proliferate without coordination.

\noindent\textbf{Institutional and repository barriers.} Willing researchers may struggle to deposit FAIR²-compliant datasets if their repositories do not support required metadata fields. While platforms like HuggingFace Hub and Zenodo accommodate flexible metadata, traditional ecological repositories may require technical updates. This creates a chicken-and-egg problem: repositories will not invest in updates without demand, but researchers will not be able adopt without repository support. Building relationships with repository managers will be critical.

\subsection{Future Work}
To address these adoption challenges, we will develop an open-source supporting library including automated compliance checkers, template generators for common drone platforms, and visualization tools for metadata exploration. Similar efforts in animal movement ecology, such as the Movement analysis package \cite{movement2024}, demonstrate how well-designed tooling can drive standards adoption by making compliance easier than non-compliance.
We will release dataset conversion scripts demonstrating FAIR² implementation across diverse use cases (habitat mapping, behavioral ecology, multi-modal coordination, landscape tracking, marine geolocation). These examples provide templates researchers can adapt and demonstrate to repositories and funding agencies that the standard addresses real research needs. We are particularly interested in connecting with repository managers to co-develop integration pathways.

Future technical extensions include autonomous fleet and mission-level provenance tracking as drones increasingly operate in coordinated swarms~\cite{rolland2025drone}, multimodal integration schemas for acoustics and eDNA, and digital-twin interoperability to link field observations with simulation environments. As drone missions become increasingly autonomous with automated area coverage and real-time computer vision tracking \cite{kline2025studying}, the need for machine-actionable metadata grows more urgent.

Standardization is an act of scientific stewardship, ensuring that data collected at great effort are findable, accessible, interoperable, reusable, and AI-ready for the benefit of all. While challenges remain, the alternative (continued fragmentation and duplicated reformatting effort) wastes both research time and scientific potential. By building on established standards, documenting realistic time investments, providing tiered compliance options, and developing practical tooling, we aim to make FAIR² adoption the path of least resistance for drone ecologists seeking to maximize their data's long-term impact.

\section*{Data and Code Availability}
Reference datasets demonstrating the standard are available at \href{https://huggingface.co/collections/imageomics/fair2-drones}{HuggingFace}. 
All validation and conversion tools are available as open-source software on GitHub at \href{https://imageomics.github.io/fair_drones/}{imageomics.github.io/fair\_drones}.

\section*{Acknowledgments}
\small
This work was supported by the Imageomics Institute, which is funded by the US National Science Foundation's Harnessing the Data Revolution (HDR) program under Award \# 2118240 (Imageomics: A New Frontier of Biological Information Powered by Knowledge-Guided Machine Learning), and the AI Institute for Intelligent Cyberinfrastructure with Computational Learning in the Environment (ICICLE) (NSF grant OAC-2112606). This work is further supported by the grant for AI-ready Ecology and Biodiversity Data Infrastructure for Open Science (NSF Award No. 2531922). This work is also supported by the AI and Biodiversity Change (ABC) Global Center. The ABC Global Center is funded by the US NSF under Award No. 2330423 and the Natural Sciences and Engineering Research Council of Canada under Award No. 585136. This dataset draws on research supported by the Social Sciences and Humanities Research Council.

This research has been carried out as part of the project WildDrone, funded by the European Union's Horizon Europe Research and Innovation Program under the Marie Skłodowska-Curie Grant Agreement No. 101071224. The WildDrone project has also received funding in part from the EPSRC funded Autonomous Drones for Nature Conservation Missions grant (EP/X029077/1). B.R.C. acknowledges support from the Deutsche Forschungsgemeinschaft (DFG, German Research  Foundation) under Germany's Excellence Strategy—‘Centre for the  Advanced Study of Collective Behaviour’ EXC 2117-422037984 and the University of Konstanz's Investment Grant program.

Any opinions, findings and conclusions or recommendations expressed in this material are those of the author(s) and do not necessarily reflect the views of the National Science Foundation, Natural Sciences and Engineering Research Council of Canada, or Social Sciences and Humanities Research Council. Views and opinions expressed are those of the author(s) only and do not necessarily reflect those of the European Union or the European Commission. Neither the EU nor the EC can be held responsible for them.

\bibliography{ref}





\end{document}